\documentclass[runningheads]{llncs}

\usepackage{multirow}%
\usepackage{graphicx}
\usepackage{amsmath,amssymb,amsfonts}
\usepackage[title]{appendix}%
\usepackage{xcolor}%
\usepackage{textcomp}%
\usepackage{manyfoot}%
\usepackage{booktabs}%
\usepackage{algorithm}%
\usepackage{algorithmicx}%
\usepackage{algpseudocode}%
\usepackage{listings}%

\begin{document}

\title{Can We Transfer Noise Patterns? A Multi-environment Spectrum Analysis Model Using Generated Cases}

\author{Haiwen Du\inst{1,2} \and
Zheng Ju\inst{2} \and
Yu An\inst{2}\and
Honghui Du\inst{2} \and
Dongjie Zhu\inst{3}*\and\\
Zhaoshuo Tian\inst{1}\and
Aonghus Lawlor\inst{2}\and
Ruihai Dong\inst{2}*}

\authorrunning{H. Du et al.}
\titlerunning{An Adaptive Denoising Model for Multi-environment Spectrum Analysis}
\institute{School of Astronautics, Harbin Institute of Technology, Harbin, China \and
Insight Centre for Data Analytics, Dublin, Ireland \and
School of Computer Science and Technology, Harbin Institute of Technology, Weihai, China\\
\email{zhudongjie@hit.edu.cn, ruihai.dong@ucd.ie}}

\maketitle

\begin{abstract}
Spectrum analysis systems in online water quality testing are designed to detect types and concentrations of pollutants and enable regulatory agencies to respond promptly to pollution incidents. However, spectral data-based testing devices suffer from complex noise patterns when deployed in non-laboratory environments. To make the analysis model applicable to more environments, we propose a noise patterns transferring model, which takes the spectrum of standard water samples in different environments as cases and learns the differences in their noise patterns, thus enabling noise patterns to transfer to unknown samples. Unfortunately, the inevitable sample-level \textit{baseline noise} makes the model unable to obtain the paired data that only differ in dataset-level \textit{environmental noise}. To address the problem, we generate a sample-to-sample case-base to exclude the interference of sample-level noise on dataset-level noise learning, enhancing the system's learning performance. Experiments on spectral data with different background noises demonstrate the good noise-transferring ability of the proposed method against baseline systems ranging from wavelet denoising, deep neural networks, and generative models. From this research, we posit that our method can enhance the performance of DL models by generating high-quality cases. The source code is made publicly available online at https://github.com/Magnomic/CNST.
\end{abstract}



\section{Introduction}\label{sec1}

Laser-induced fluorescence spectrum (LIFS) analysis is used to interpret spectrum data obtained via laser-induced fluorescence (LIF) spectroscopy in a specific range of wavelengths, frequencies, and energy levels \cite{hu2019selection}. Its advantages include the high sampling frequency, high accuracy, and easy deployment, making it a popular water quality monitoring approach \cite{zacharioudaki2022review,bukin2020new}. However, avoiding the impact of the noise on the analysis results is challenging for multi-environment deployment because the noise is caused by changes in weather, temperature, or water impurities \cite{laurent2019denoising,yang2021miniaturization}. 

Denoising is essential in analysing spectral and other one-dimensional (1D) data, as it can improve the accuracy and reliability of subsequent analyses. Traditional methods that rely on mathematical models or filters (e.g., moving average or wavelet transformation) can reduce high-frequency or specific noise patterns while consuming less computing resources \cite{sobolev2013application}. However, they cannot learn complex nonlinear relationships between the noisy and the corresponding clean signal. Deep learning(DL)-based denoising methods use a data-driven approach. They are trained on a large dataset to learn the underlying patterns and relationships between the data \cite{chen2022olive,kazemzadeh2022deep}. 

Nevertheless, we found limitations in this approach due to the complex noise sources and noise patterns in LIFS. Specifically, we cannot obtain paired clean-noisy data to analyse the noise pattern between two datasets because the noise in spectral data is from both \textit{baseline noise} $\xi$ and \textit{environmental noise} $N$ \cite{abdessamad2014spectrum,santos2004effects}. \textit{Environmental noise} arises from backgrounds, light sources and other stable factors over time, exhibiting differences at the dataset-level. The \textit{baseline noise} originates from random disturbances during the sampling process, such as baseline drift, which is present even in laboratory environments. Different from \textit{environmental noise}, \textit{baseline noise} shows differences at the sample-level, i.e., every sample has different \textit{baseline noise}. However, as both noise sources are low-frequency and superimposed on the signal, extracting only the \textit{environmental noise} from the LIFS to train a noise patterns transferring model is complex. We can only feed the samples with both \textit{environmental noise} and \textit{baseline noise} differences to the noisy pattern learning model, which is inefficient.

In this paper, we propose a method to enhance the performance of denoising models based on generated cases, i.e., we generate paired samples that differs only in \textit{environmental noise}. In turn, we learn the differences to achieve noise pattern transformation. Firstly, the model extracts noise patterns from a dataset-to-dataset (D2D) case-base that consists of two groups of samples from dataset $D_\mathbb{T}$ and dataset $D_\mathbb{S}$. Then, we generate a sample-to-sample (S2S) case-base in which all pairs of the sample differ only in going from \textit{environmental noise} $N_\mathbb{S}$ to $N_\mathbb{T}$ ($N_\mathbb{S-T}$). Finally, we train a noise patterns transferring model using the S2S case-base. The highlights are:

1. we develop a noise transfer model to transfer the \textit{environmental noise} pattern  $N_\mathbb{S}$ of dataset $D_\mathbb{S}$ into the \textit{environmental noise} pattern $N_\mathbb{T}$ of dataset $D_\mathbb{T}$ (also denoising when $D_\mathbb{T}$ is \textit{environmental noise}-free);

2. propose a network structure for extracting \textit{environmental noise} $N_\mathbb{S}$, signal $X$, and \textit{baseline noise} $\xi_{T}$ from a D2D case $S$ and $T$, thereby generating a new sample $G$ that only have \textit{environmental noise} pattern differences $N_\mathbb{S-T}$ with sample $T$. It is the first work to enhance the learning performance of denoising models by generating new cases; and

3. verify the contribution of our model on 1D LIFS noise patterns transferring tasks. It significantly improves the performance of chemical oxygen demand (COD) parameter analysis model under different noise patterns.

In the following section, we discuss relevant related work and present motivations for this work. Section \ref{sec3} presents the proposed model that captures noise pattern differences between signals in a different dataset. In Section \ref{sec4}, we describe the experimental setup. Then, Section \ref{sec5} presents and analyses the evaluation results. Section \ref{sec6} concludes the paper and discusses future work.
\section{Background and Motivation}\label{sec2}


\subsection{Problems in Traditional Denoising Models}\label{subsec21}

In LIF spectroscopy, noise can arise from various sources, including electronic noise in the sensors, light source fluctuations, and sample matrix variations. It can affect the LIFS analysis process by introducing artefacts, such as spurious peaks or baseline drift, that can obscure or distort spectral features associated with water quality parameters \cite{wang2019independent}. For example, baseline drift is a noise that can make it challenging to identify and quantify the fluorescence peaks accurately. Similarly, high-frequency noise can introduce spurious peaks in the spectra that can be misinterpreted as actual spectral features \cite{loh2020operation}. Therefore, avoiding the noise's influence on the analysis result is vital for researchers in analysing LIFS.

Various denoising techniques can mitigate noise in LIFS, such as smoothing algorithms or wavelet transform-based methods \cite{sobolev2013application}. These methods can remove parts of noise while preserving the spectral features of interest. Another popular approaches are machine learning algorithms, such as DNNs \cite{hu2019identification}. Compared with traditional methods, their adaptability to various noise types and their robustness to complex scenes make DNNs an attractive option for denoising tasks.

\begin{figure}
    \centering
    \includegraphics[width=1.0\textwidth,trim=40 80 20 90,clip]{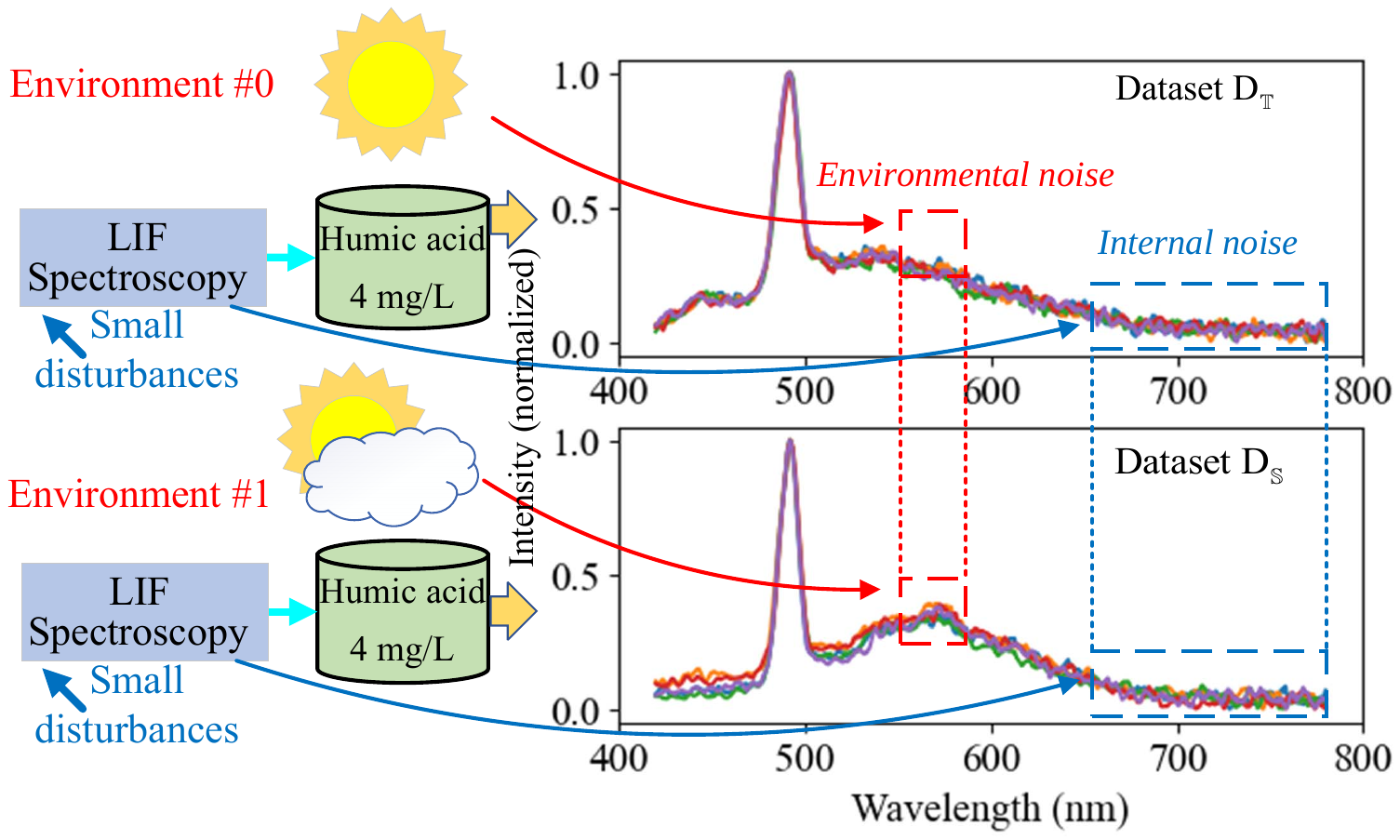}
    \caption{We provide LIFS examples of two standard solutions in different environments. The blue boxes show that each sample has a different \textit{internal noise} $\xi$, even from the same dataset. The red boxes show the \textit{environmental noise} $N$, at the dataset level, are shared by all the samples in the same dataset, respectively. For ease of presentation, all $S$ and $T$ that occur together in the following paper represent samples corresponding to the same water sample.}\label{fig0}
\end{figure}

Nevertheless, we are facing a dilemma in applying DNNs, which is caused by the \textit{internal noise}, i.e., even in the laboratory environment, the spectral data still carries low-frequency noise because of fuzzy factors such as baseline drift. To make matters worse, the two measurements for the same water sample show different \textit{internal noise} patterns. For datasets $D_{\mathbb{S}}$ and dataset $D_{\mathbb{T}}$ collected in different environments, when we try to transfer the \textit{environmental noise} pattern of a sample $S$ in dataset $D_\mathbb{S}$ into the \textit{environmental noise} pattern of dataset $D_\mathbb{T}$, which have the relations in Eq. \ref{eq6}, where $S$ and $T$ are the LIFSs of the same water sample $X$ in $D_{\mathbb{S}}$ and $D_{\mathbb{T}}$, $N_{\mathbb{S} - \mathbb{T}}$ is the \textit{environmental noise} difference (dataset-level) between $D_{\mathbb{S}}$ and $D_{\mathbb{T}}$, $\xi_{S}$ and $\xi_{T}$ are the \textit{internal noise} (sample-level) in samples $S$ and $T$.

\begin{equation}
    \begin{aligned}
        S = X + N_{\mathbb{S}} + \xi_{S}  \\ 
        T = X + N_{\mathbb{T}} + \xi_{T} \\ \label{eq6}
        N_{\mathbb{S}-\mathbb{T}} = N_{\mathbb{T}} - N_{\mathbb{S}}
    \end{aligned}
\end{equation}
We find that it does not satisfy the relationship in Eq. \ref{eq1} but in Eq. \ref{eq2}. Then, we give examples in Fig. \ref{fig0} to better represent the terms in noise patterns.
\begin{equation}
    T = S - N_{\mathbb{S}-\mathbb{T}}\label{eq1}
\end{equation} 
\begin{equation}
    T = S - N_{\mathbb{S}-\mathbb{T}} - \xi_{S} + \xi_{T} \label{eq2}
\end{equation}

Unfortunately, the DNN models show bad performance when learning the dataset-level noise pattern transformation from $N_{\mathbb{S}}$ to $N_{\mathbb{T}}$ because of the existence of sample-level noise $\xi_{S}$ and $\xi_{T}$.

\subsection{Studies Related to Noise Pattern Transferring}\label{subsec22}

Since \textit{internal noise} is inevitable in LIFS, we hope all the samples have the same \textit{environmental noise} patterns by removing their \textit{environmental noise} patterns differences. With this, we only need to train the analysis model in one \textit{environmental noise} pattern. Then we can process the data in various environments by transferring their \textit{environmental noise} to the target pattern \cite{du2022disentangling}.

We refer to the generative methods, which use the LIFS of the same standard solution in different environments as cases. Therefore, noise pattern differences can be learned by analysing these cases and applied to transfer noise patterns of unknown samples. However, directly training a DNN-based noise-transferring model like DnCNN \cite{zhang2017beyond} using these cases is less efficient because the \textit{internal noise} is sample-level low-frequency noise and superimposed on the signal and \textit{environmental noise}. 

We have two feasible approaches to achieve it. The first is to design a model that does not strictly require pairwise data for training, which CycleGAN \cite{kwon2019predicting} represents. CycleGAN uses an adversarial and cycle consistency loss, encouraging the network to learn a consistent mapping between the two domains. Recently, CycleGAN has been used to suppress unknown noise patterns and get satisfactory results in 1D signal processing, such as audio and seismic traces \cite{kaneko2019cyclegan}.

\begin{figure}
    \centering
    \includegraphics[width=1.0\textwidth,trim=60 135 30 155,clip]{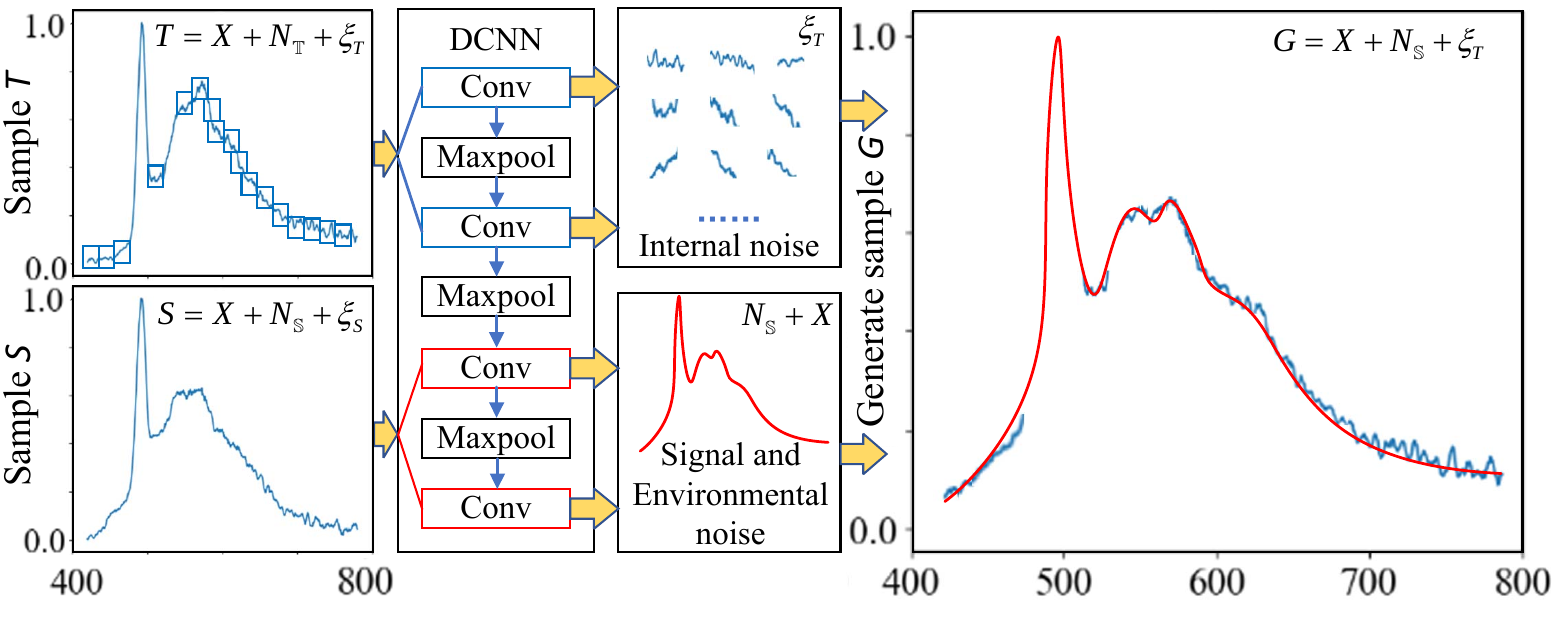}
    \caption{We provide an example of our hypothesis. Sample $S$ and sample $T$ are fed into the DCNN to get their feature maps on deep and shallow convolutional layers, which correspond to extracting \textit{internal noise} and \textit{environmental noise} (also the signal). Then, we use this information to synthesise $G$.  }\label{fig_gene}
\end{figure}

Alternatively, we create new cases that have \textit{environmental noise} (dataset-level) difference but the same \textit{internal noise} (sample-level). Compared with CycleGAN, it is more explainable and provides a much more guided process. It is challenging to extract only \textit{environmental noise} because \textit{internal noise} is superimposed on it. Although both \textit{internal noise} and \textit{environmental noise} are low-frequency noise, the frequency of the \textit{internal noise} is slightly higher than the \textit{environmental noise}. It means that the curve fluctuations caused by internal noise are in the smaller wavelength range. As the feature maps in deep convolutional neural network (DCNN) can help to generate cases \cite{kenny2021generating} and feature information \cite{turner2019nod}, we try to use a pre-trained DCNN to extract these noise patterns separately. Based on the theory that the deeper the convolutional layer is, the more pooling layers it passes through and the increasing range of its extracted features\cite{turner2018novel}, we propose the hypothesis that deeper convolutional layers can better extract the \textit{environmental noise} $N_{\mathbb{S}}$ and signal $X$ of the samples in datasets $D_{\mathbb{S}}$, and shallower convolutional layers can better extract the \textit{internal noise} $\xi_{T}$ of the samples in datasets $D_{\mathbb{T}}$. We can use them and generate a sample $G = X + N_{\mathbb{S}} + \xi_{T}$ such that its difference from $T$ is $N_{\mathbb{S} - \mathbb{T}}$, as shown in Fig. \ref{fig_gene}.

\section{Noise Patterns Transferring Model}\label{sec3}

\begin{figure}
    \centering
    \includegraphics[width=1.0\textwidth,trim=155 120 155 120,clip]{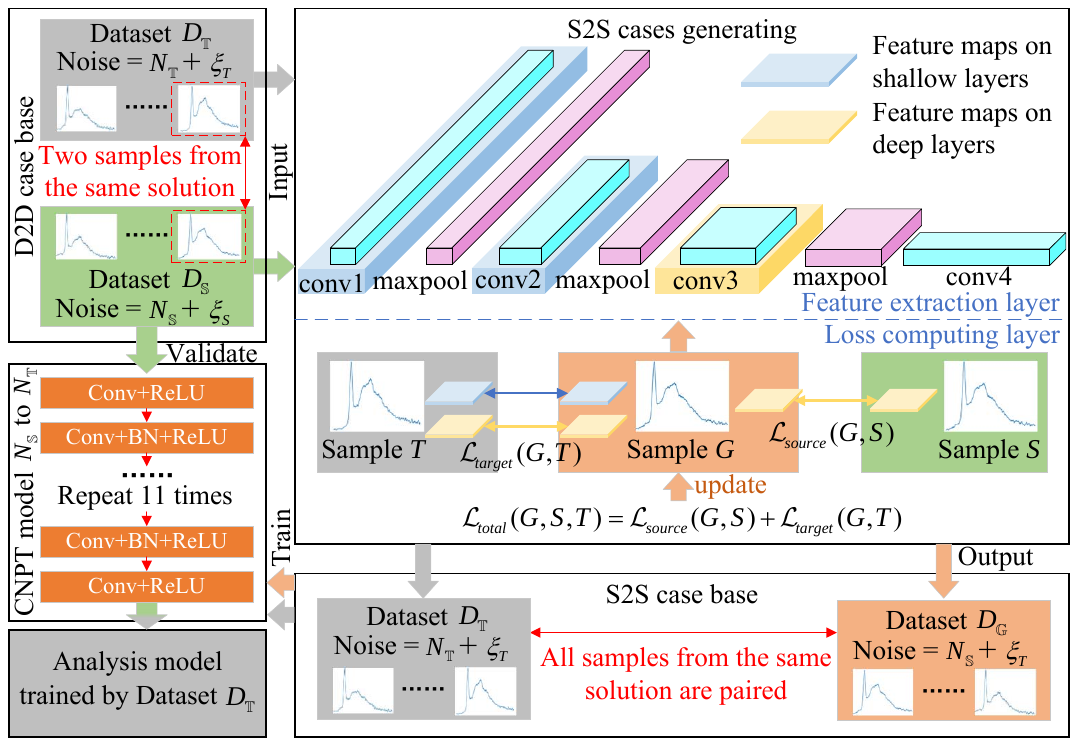}
    \caption{The workflow of our method. $D_\mathbb{S}$ and $D_\mathbb{T}$ are collected in different environments. Firstly, we randomly select a case from D2D case-base, i.e., $S$ and $T$ and feed them with an initialised sample $G$ into the pre-trained 1D DCNN model for extracting their feature maps. Secondly, we get the generated $G$ by minimising the loss using the feature maps of $G$, $S$ and $T$. Then we get an S2S case that is consisted of $G$ and $T$, which with the same $\xi_{T}$ and only differ in $N_{\mathbb{S}}$ and $N_{\mathbb{T}}$. Finally, we use all of the generated S2S cases, i.e., S2S case-base, to train the curve noise patterns transferring (CNPT) model, which can transfer the noise patterns of samples from $N_{\mathbb{S}}$ to $N_{\mathbb{T}}$. }\label{fig1}
\end{figure}

Since the samples in different datasets have both sample-level (noise difference in \textit{internal noise}) and dataset-level (noise difference in \textit{environmental noise}) noise pattern differences, traditional solutions meet difficulties when trying to extract and learn \textit{environmental noise} without the interference of \textit{internal noise}. This section details our generated cases (GC)-based method to achieve noise patterns transferring. The overall workflow of our method is shown in Fig. \ref{fig1}.

\subsection{D2D Case-base and S2S Case-base}\label{subsec31}

Our system has two case-bases, D2D case-base and S2S case-base, if we want to transfer the \textit{environmental noise} pattern $N_{\mathbb{S}}$ in $D_{\mathbb{S}}$ to the \textit{environmental noise} pattern $N_{\mathbb{T}}$ in $D_{\mathbb{T}}$, a D2D case is a sample $S$ in $D_{\mathbb{S}}$ and random sample $T$ in $D_{\mathbb{T}}$ that is measured using the same standard solution but under different environments. Although all of the S2S and D2D cases have the same $N_{\mathbb{S}-\mathbb{T}}$, D2D cases have different \textit{internal noise} while S2S cases have the same \textit{internal noise}. In other words, an S2S case is a pair of a sample $G$ in $D_{\mathbb{G}}$ and a sample $T$ that has the same \textit{internal noise} $\xi_{T}$ but differs in $N_{\mathbb{S}}$ and $N_{\mathbb{T}}$. 

Since S2S cases cannot be directly obtained by LIF spectroscopy because of the existence of \textit{internal noise}, to learn the pattern of $N_{\mathbb{S}-\mathbb{T}}$, the first step is to generate S2S cases using the D2D case-base that satisfies Eq. \ref{eq7}.
\begin{equation}
        G = T + N_{\mathbb{S}-\mathbb{T}} \label{eq7}
\end{equation}

According to Eq. \ref{eq6}, $G$ can be represented as Eq. \ref{eq9}.
\begin{equation}
    \begin{aligned}
    G &= S - \xi_{S} + \xi_{T} \\ 
      &= X + N_{\mathbb{S}} + \xi_{T} \label{eq9}
    \end{aligned}
\end{equation}

To this step, we get a S2S case, i.e., $G$ and $T$, because they only have the difference in $N_{\mathbb{T}}$ and $N_{\mathbb{S}}$. It enables us to train a model that can learn the pattern of this transformation. The key point, therefore, is how to extract $X$, $N_{\mathbb{S}}$ and $\xi_{T}$ from D2D case-base. 

\subsection{S2S Cases Generating Model}\label{subsec32}

According to the hypothesis in section \ref{subsec22}, we propose a `feature extraction-matching' method to generate $G$ in the S2S case. Firstly, we use a pre-trained DCNN-based 1D LIFS analysis model that contains multiple convolutional and pooling layers to extract features. Then, we initialise $G$ and design a loss function that makes the feature of $G$ approximate $X$, $\xi_{T}$, and $N_{\mathbb{S}}$ from the D2D case. Lastly, we get an S2S case that is consisted of $G$ and $T$ by minimising the loss.

As $X$, $N_{\mathbb{T}}$ and $\xi_{T}$ are wavelength (x-axis) related but show different signal frequency features. We propose a loss function $\mathcal{L}_{total}$ that consists of source loss $\mathcal{L}_{s}$ and target loss $\mathcal{L}_{t}$ in Eq. \ref{eq10}, where $\alpha$ and $\beta$ are the weights. $l_t$ and $l_s$ are the layers to compute $\mathcal{L}_{t}$ and $\mathcal{L}_{s}$.
\begin{equation}
    \mathcal{L}_{total}(G,S,T,l_s,l_t) = \alpha\mathcal{L}_{s}(G, S, l_s) + \beta\mathcal{L}_{t}(G, T, l_s, l_t)\label{eq10}
\end{equation}

$\mathcal{L}_{s}$ is used to construct the $N_{\mathbb{S}}$ and $X$ in the generated sample $G$. As $N_{\mathbb{S}}$ and $X$ are x-axis related and show lower-frequency features, we use deeper feature maps and position-related loss to reconstruct $X$ and $N_{\mathbb{S}}$ from $S$ (see Eq. \ref{eq11}.)

\begin{equation}
    \mathcal {L}_{s}\left ({G, S, l_s}\right) =  \left ({F_{i}^{l_{s}}\left ({G}\right) - F_{i}^{l_{s}}\left ({S}\right) }\right)^{2}\label{eq11}
\end{equation}

\begin{figure*}
    \centering
    \includegraphics[width=1.0\textwidth,trim=190 260 295 200,clip]{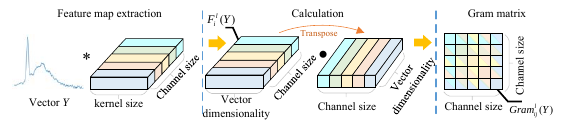}
    \caption{shows how the gram matrix is calculated. Suppose that we input the sample $Y$ to the DCNN for its convolution layer $l$; we get the feature matrix with the size of $P_{l} * M_{l}$, where $P_{l}$ is the channel size of $l$, $M_{l}$ is the size of feature map in $l$, the convolution of $Y$ on the $i^{th}$ channel is $F_{i}^{l}(Y)$. Then, by multiplying the obtained matrix with its transpose, we obtain the gram matrix of $P_{l} * P_{l}$, where $Gram_{i,j}^{l}(Y)$ represents the correlation of $Y$ on the $i^{th}$ and $j^{th}$ features in convolution layer $l$.}\label{fig2}
\end{figure*}

$\mathcal{L}_{t}$ is used to construct the \textit{internal noise} $\xi_{T}$ of $T$ in $G$. As we discussed, its frequency is higher than \textit{environmental noise} and signal. We use feature maps on shallow layer $l_t$ and gram matrix (see Eq. \ref{eq12} and Fig. \ref{fig2}) to control its feature shape and an MSE loss on deep layer feature $l_s$ to control its x-axis position. The reason we introduced the gram matrix is that we cannot synthesise $\xi_{T}$ with the same loss function structure for $T$ as for $S$. It would cause the overall loss to converge towards $\mathcal {L}_{s}$ or $\mathcal {L}_{t}$, and thus we would not be able to obtain features from both $S$ and $T$.

\begin{equation}
Gram_{ij}^{l_t}\left ({Y}\right) = \sum _{k \in l_t} F_{ik}^{l_t}\left ({Y}\right) \cdot F_{jk}^{l_t}\left ({Y}\right)\label{eq12}
\end{equation}

The overall $\mathcal{L}_{t}$ is calculated as Eq. \ref{eq13}, where $\omega _{l}$ is the loss weight of $l$.
\begin{equation}
    \begin{aligned}
        \mathcal {L}_{t}\left (G, T, l_s, l_t\right)=&\sum _{i} \left ({F_{i}^{l_{s}}\left ({G}\right) - F_{i}^{l_{s}}\left ({T}\right) }\right)^{2} + \sum _{l \in l_{t}}\omega _{l}E_{l} \\
        E_{l}=&\frac {1}{4P_{l}^{2}M_{l}^{2}}\sum _{i,j} \left ({Gram_{ij}^{l}\left (G\right)-Gram_{ij}^{l}\left (T\right) }\right) ^{2} \label{eq13}
    \end{aligned}
\end{equation}
\subsection{Noise Patterns Transferring Model}\label{subsec33}

S2S case-base provides noise pattern differences $N_{\mathbb{S}-\mathbb{T}}$ in sample-to-sample pairs, i.e., $G$ and $T$. It enables us to train the CNPT model with $CNPT(G) = T$. Similar to denoising tasks, we use the residual learning method, which involves learning the residual mapping between the pairwise samples to obtain the CNPT model that achieves transferring $N_{\mathbb{S}}$ to $N_{\mathbb{T}}$.

The CNPT model's network structure consists of 17 convolutional layers and rectified linear unit (ReLU) layers, allowing it to capture complex patterns and features. The inputs to CNPT are the generated sample $G = X  + N_{\mathbb{S}}+ \xi_{T}$ and the corresponding sample $T = X + N_{\mathbb{T}} + \xi_{T}$. CNPT is trained to learn the residual mapping between the S2S case $\mathcal {R}(G, \phi) = N_{\mathbb{S}-\mathbb{T}} = N_{\mathbb{T}} - N_{\mathbb{S}}$.

We use the MSE loss function to train the CNPT to minimise the learned noise pattern residual and the expected residuals. The loss function is shown in Eq. \ref{eq14}, where the $Q$ represents the number of S2S cases we feed the model.
\begin{equation}
    \mathcal {L}_{CNPT} =\frac {1}{2Q} \sum _{i=1}^{Q}\left \|{\mathcal {R}({G}; \phi)-\left ({G - T}\right)}\right \|^{2}\label{eq14}
\end{equation}

Ideally, the trained CNPT model should learn the noise pattern differences between $N_{\mathbb{T}}$ and $N_{\mathbb{S}}$ by feeding S2S cases into it.

\section{Environmental Setup}\label{sec4}

To verify the performance of our GC-based noise pattern transferring model, we try to reproduce actual deployment scenarios in experiments. Therefore, samples measured in different environments, i.e., with different environmental noises, will be used in the verification process. We use the task to measure the COD parameter using LIFS, which can estimate the amount of organic matter in a water sample.

In this section, we introduce how our model is deployed to fit the environmental changes and the configuration of our model. Further, the datasets and the baselines are also provided in detail.

\subsection{Deployment settings and runtime environments}\label{subsec41}

\begin{figure}
    \centering
    \includegraphics[width=1.0\textwidth,trim=60 140 60 130,clip]{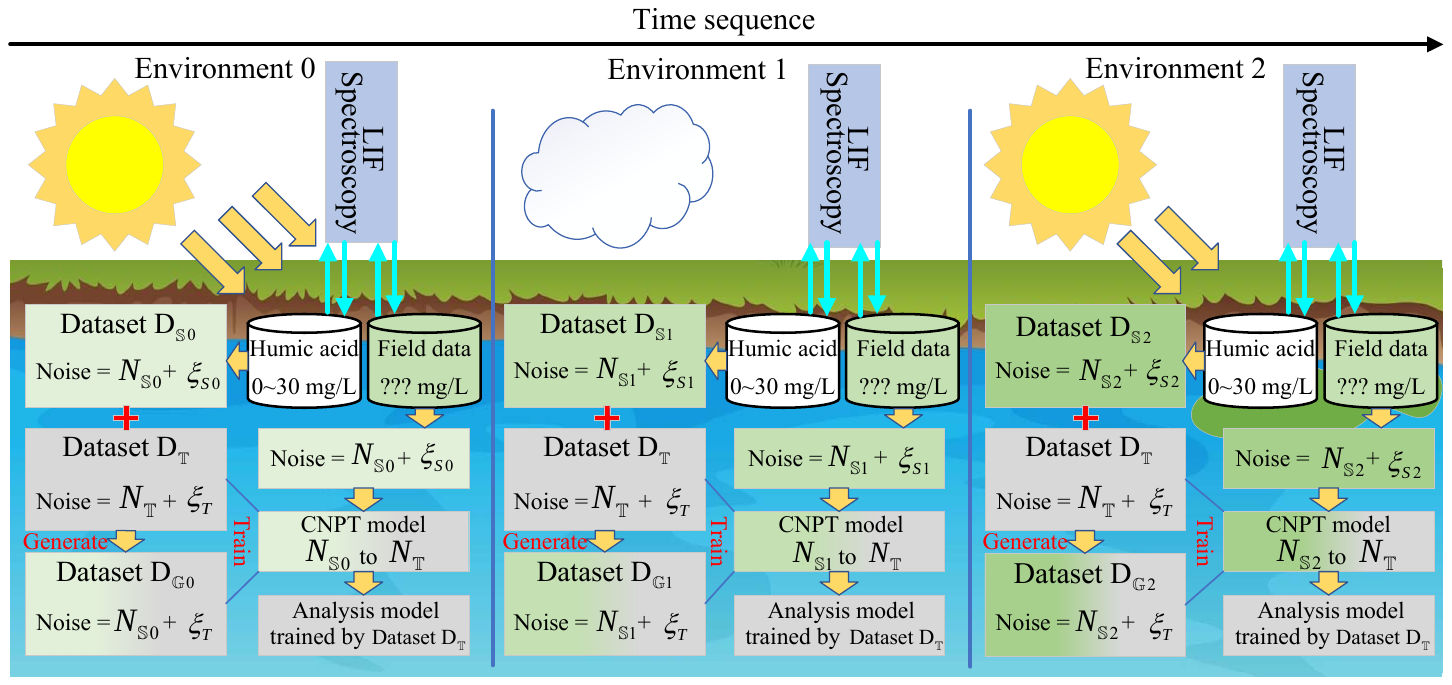}
    \caption{We show three environments in the figure: strong sunlight, weak sunlight, and static pollutant background, which has different \textit{environmental noise}. The system can periodically test the LIFS of standard water samples as the environment changes, correcting the CNPT model by retraining to process the new \textit{environmental noise}.}\label{fig3}
\end{figure}

Unlike traditional solutions, our approach can process new noise patterns by continuously collecting standard spectral samples from the environment. The workflow is shown in Fig. \ref{fig3}.

The standard water samples are humic acid solutions that can help get accurate COD parameters of water by enhancing the oxidation process and reducing interference from other substances. To ensure the CNPT model can handle samples of unknown COD concentration, we use parts of standard water sample groups to generate the S2S case-base and train the CNPT model. We prepared three environments: darkroom, static natural lighting, and static pollutant background. All the experiments were performed on a Linux server for the hardware environments, with CUDA version 11.8 using GeForce GTX 4090 Graphics Cards. We use the PyTorch framework to write the codes. 

\subsection{Model configurations}\label{subsec42}

In this section, we introduce all the hyper-parameters and the configurations in the experiments to guide the researchers in refining the model in different applications and environments. 

For the S2S case generating model, the feature extraction model is a four-layer 1D convolutional neural network in which kernel size is 1x7. This pre-trained model is from a COD parameter analysis model, which is trained by 6,000 samples and achieves 92.53\% accuracy. The feature maps for calculating $\mathcal{L}_{t}$ are from conv1 and conv2 layers, while conv3 is used for x-axis-related MSE loss. The ratio of $\alpha$ and $\beta$ is set to 1: 2e5 and $\omega_{l} = 0.2 * \omega_{l-1}$. We iteratively execute the L-BFGS function 150 times to minimise the loss.

For the CNPT model, the kernel size is 1x7. The training batch size is 64, and the training epochs are 100. The learning rate we use is 1e-3, and the input data is normalised by the min-max normalisation method before feeding into the neural network. The data for the training-to-validation ratio is 80\% to 20\%.

The COD parameter analysis model, used for verifying the performance of noise patterns transferring, has the same network structure as the feature extraction model. The training batch size is 128, the epochs are 200, and the learning rate is 3e-4. All the data fed to this model for validation are not used for training or testing the CNPT model.

\subsection{Datasets}\label{subsec43}

\begin{figure}
    \centering
    \includegraphics[width=1.0\textwidth,trim=0 0 0 0,clip]{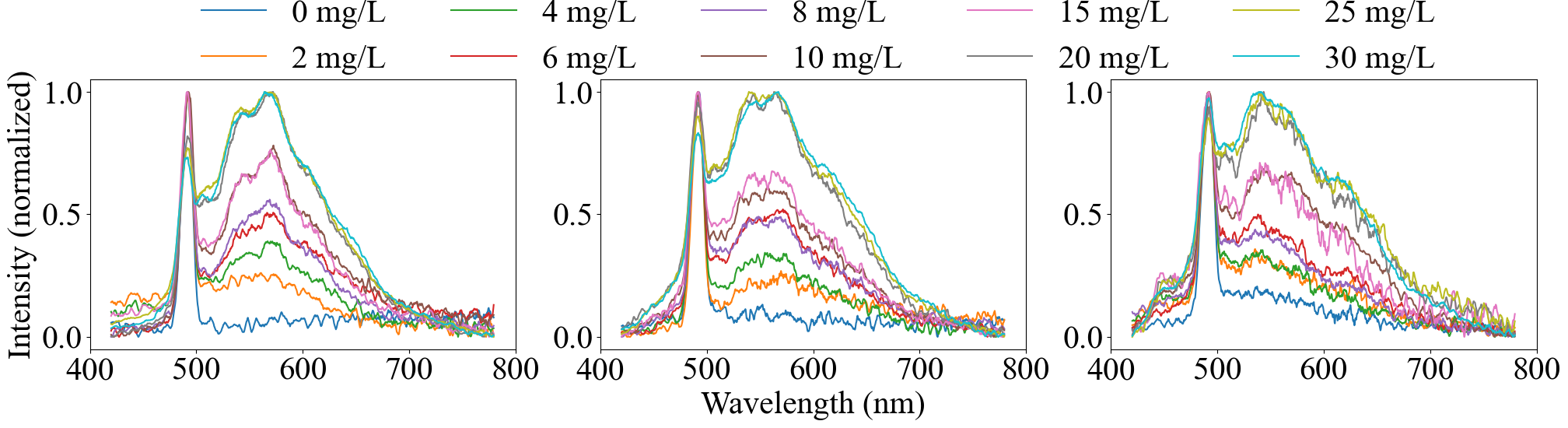}
    \caption{The LIFS of standard water sample under three noise patterns: The left figure represents a dark room setting ($D_\mathbb{A}$), the central one portrays a natural light condition ($D_\mathbb{B}$), while the right one is from a static pollutant background environment ($D_\mathbb{C}$).}
    
\label{fig4}
\end{figure}

The datasets are LIF spectral data of humic acid solution with different noise patterns, open-sourced at \cite{7r0c-mf67-23}. The COD of solutions is measured using laser-induced fluorescence spectroscopy \cite{tian2019rapid}, in which a 405 nm semiconductor laser was used as the excitation source. The excited light signals from the water sample pass through a spectroscopic filter system and be focused into the CCD by a lens. The dataset contains three groups of LIFS, which are collected from 10 solutions that the COD parameter ranges from 0 mg/L to 30 mg/L in three different noise patterns $D_\mathbb{A}$, $D_\mathbb{B}$, and $D_\mathbb{C}$. The noise strength of them is ordered as $D_\mathbb{A} < D_\mathbb{B} << D_\mathbb{C}$. Each solution sample is measured 200 times, and all the data are min-max normalised. We provide examples of LIF spectra in Fig. \ref{fig4}.

\subsection{Baselines}\label{subsec44}

To verify the performance of our model, we select competitive models with good performance in terms of noise processing as baselines. In other words, they are used for transferring noise patterns, not to get spectral analysis results. All of our model's and baselines' outputs are fed into the same COD parameter analysis model that is trained by the samples in $D_\mathbb{T}$, which can check whether the samples that are transferred to the noise pattern $N_{\mathbb{T}}$ can be correctly identified by the analytical model trained with $D_\mathbb{T}$. The baselines are as follows:

\textbf{Rule-based Models}: use wavelets, Fourier transforms, or filtering to reduce the noise of the data. They have the advantage of not requiring paired data and learning processes. For LIFS, we use the `db8' wavelet to achieve the denoising task\cite{he2013tunable}. This method trains the classification model using the denoised samples in $D_\mathbb{T}$. Then, the CNPT model learns the noise pattern difference between denoised samples in $D_\mathbb{S}$ and $D_\mathbb{T}$.

\textbf{DNN Models}: use end-to-end training process. DnCNN \cite{zhang2017beyond} is a state-of-the-art image denoising method that uses residual learning to remove noise from images. In this paper, we implemented a 1D DnCNN model with a 1x7 kernel size to learn the noise pattern differences between $D_\mathbb{S}$ and $D_\mathbb{T}$. Since the spectral data in different noise patterns are not paired, we randomly pick samples in $D_\mathbb{S}$ and $D_\mathbb{T}$ corresponding to the same standard solution as paired data to train it.

\textbf{Generative Models}: learn at the scale of the dataset, which makes the model find noise pattern differences in the adversarial learning process. CycleGAN \cite{kwon2019predicting} is a deep learning model that can learn to translate images from one domain to another without paired examples. Therefore, CycleGAN is a suitable approach to handle the unpaired LIFS data. We implemented a 1D CycleGAN model, which can transfer the noise pattern of samples from $D_\mathbb{S}$ to $D_\mathbb{T}$.

\section{Experimental Results}\label{sec5}

We discuss the experimental results in two parts: 1) check if the gram matrix works well in the S2S cases generating process; 2) test the model's performance by feeding the output of CNPT to a trained COD parameter analysis model.

\subsection{Effects on S2S cases generating}\label{subsec51}

\begin{figure}
    \centering
    \includegraphics[width=1.0\textwidth,trim=180 58 190 57,clip]{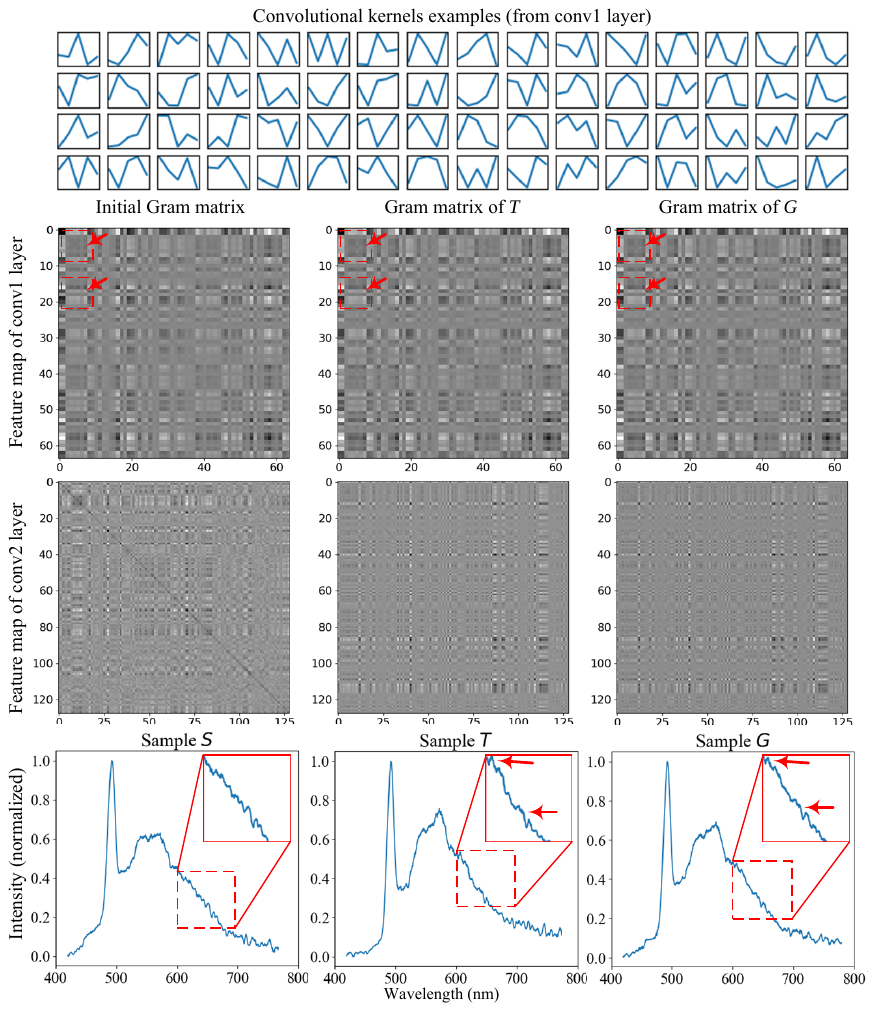}
    \caption{We show the convolution kernel used to compute the feature map in the top part of the figure. Then, we show the initial, target, and final state of the gram matrix of the signal $G$ in the training progress of the conv1 and conv2 layers. The bottom part shows the corresponding signals $T$, $S$, and $G$ in this example.}\label{fig5}
\end{figure}

One of the contributions of our method is to use the unpaired data with different \textit{environmental noise} patterns and \textit{internal noise} to generate paired data that only have \textit{environmental noise} differences. It comes from our S2S cases generation process based on the `feature extraction-matching' method and the loss on different feature maps. Therefore, this section shows how the sample $G$ is generated in feature maps and gram matrix view. We present the convolution kernel and the gram matrix changes in the training process in Fig. \ref{fig5}. 

Since the convolution kernel provides rich patterns for the feature maps, we can also analyse the signal patterns from the feature maps. For example, we find that the feature in which id is 0 has a strong auto-correlation on the conv1 (the value at position 0,0 in the gram matrix), which means the pattern corresponding to the convolution kernel with ID 0 (upper left corner in the convolution kernel example) appears most frequently in the signal. 

It matches our expectation since its pattern is similar to the baseline drift, i.e., the initialisation signals and $T$ have large oscillations on a smaller scale (1x5). In contrast, conv2 has a more extensive feature extraction scale (1x25) than conv1, showing more differences in the feature map between the initialised signal and $T$ caused by the different noise patterns. From the results, the $G$ we obtained is similar to that of $T$ in \textit{internal noise} and both layers of feature maps, which confirms that our method can extract and synthesise $G$ with good quality.

\subsection{Performance on analysis model}\label{subsec52}

The most intuitive way to evaluate the noise patterns transferring models is to analyse their output using a model trained by the samples with the target noise pattern. In this section, we validate the proposed method in comparison with baselines.

The baselines we use are wavelet denoising \cite{he2013tunable}, autoencoder (AE) \cite{chandra2014adaptive}, 1D DnCNN \cite{liu2022distributed}, and 1D CycleGAN \cite{kaneko2017sequence}. They are representatives of traditional, DNN, and adversarial generation methods and are suitable for the target task. Besides, we present the accuracy of the COD parameter analysis model using the unprocessed data in $D_\mathbb{S}$ and $D_\mathbb{T}$, which provides references to the performance of baselines. 

We use two training ratios (30\% and 50\%) to train the noise transfer model, i.e., use only a training set in 3 and 5 of 10 standard water solutions, while all ten groups of data are used for the validation set. It can indicate the generalisation ability and the effect of the noise patterns transferring method with generated cases (GC). The experimental results are shown in Table. \ref{tab1}.

\begin{table}
    \caption{The experimental results of our model and baselines. Transfer targets correspond to $D_\mathbb{S}$ and $D_\mathbb{T}$. For example, when the transfer target is $N_\mathbb{B}$ to $N_\mathbb{A}$, it means $D_\mathbb{S} = D_\mathbb{B}$ and $D_\mathbb{T} = D_\mathbb{A}$ and the analysis model is trained by $D_\mathbb{A}$. For the results, $D_\mathbb{S}$ and $D_\mathbb{T}$ show the accuracy when feeding the unprocessed data in $D_\mathbb{S}$ and $D_\mathbb{T}$ to the analysis model. In contrast, others feed the data in $D_\mathbb{S}$ that are processed by corresponding  baselines to the analysis model. Each group of results uses two training ratios: 30\% and 50\%, which indicates how many standard water solutions we use when training the CNPT model. We also tested the results using an autoencoder (AE) as a CNPT model, which tests if our generated cases can improve the performance of different case-based noise transfer DL models.}\label{tab1}
    \centering
    \begin{tabular}{@{}ccccc|cc|cc|c@{}}
    \hline
    Transfer&Traning&\multirow{2}{*}{$D_\mathbb{S}$}&\multirow{2}{*}{Wavelet}&\multirow{2}{*}{CycleGAN}\;&\multirow{2}{*}{AE}&\multirow{2}{*}{GC-AE}&\multirow{2}{*}{DnCNN}&GC&\multirow{2}{*}{\;\;$D_\mathbb{T}$}\;\;\;\\
    target&ratio&&&&&&&-DnCNN&\\
    \hline
\multirow{2}{*}{$N_\mathbb{B}$ to $N_\mathbb{A}$}&50\%&\multirow{2}{*}{63.83}&74.90&76.75&50.60&57.00&71.50&\textbf{82.60}&\multirow{2}{*}{90.78}\\
&30\%&&78.40&76.50&53.25&57.30&74.25&\textbf{82.58}&\\
\hline
\multirow{2}{*}{$N_\mathbb{C}$ to $N_\mathbb{A}$}&50\%&\multirow{2}{*}{33.94}&59.30&37.50&37.45&42.85&55.80&\textbf{73.55}&\multirow{2}{*}{90.78}\\
&30\%&&69.30&39.10&30.70&44.80&62.15&\textbf{70.85}&\\
\hline
\multirow{2}{*}{$N_\mathbb{C}$ to $N_\mathbb{B}$}&50\%&\multirow{2}{*}{55.11}&62.60&66.25&48.15&53.45&58.95&\textbf{88.60}&\multirow{2}{*}{98.78}\\
&30\%&&73.45&65.25&53.35&57.00&69.45&\textbf{85.05}&\\
\hline
\multirow{2}{*}{$N_\mathbb{A}$ to $N_\mathbb{B}$}&50\%&\multirow{2}{*}{59.67}&49.40&37.25&39.15&49.05&65.35&\textbf{94.00}&\multirow{2}{*}{98.78}\\
&30\%&&58.20&42.25&39.00&45.30&63.70&\textbf{94.95}&\\
\hline
\multirow{2}{*}{$N_\mathbb{A}$ to $N_\mathbb{C}$}&50\%&\multirow{2}{*}{29.72}&53.65&40.20&23.65&31.35&51.20&\textbf{56.25}&\multirow{2}{*}{88.28}\\
&30\%&&52.90&46.85&27.20&38.45&53.95&\textbf{56.26}&\\
\hline
\multirow{2}{*}{$N_\mathbb{B}$ to $N_\mathbb{C}$}&50\%&\multirow{2}{*}{36.44}&54.85&49.75&39.45&39.70&52.45&\textbf{59.40}&\multirow{2}{*}{88.28}\\
&30\%&&60.90&57.90&39.15&41.60&59.85&\textbf{63.80}&\\
\hline
    \end{tabular}
\end{table}

\subsection{Analysis and discussions}\label{subsec53}

The results show that GC-based DnCNN CNPT model (GC-DnCNN) performs best and has a significant advantage over the baselines. It validates that our method can generate cases with better quality, which helps CNPT model learn differences in noise patterns between datasets, thereby improving the analysis model's performance. It should be noted that GC-AE, which is trained by S2S case-base performs better than AE. It better indicates that the performance of the noise patterns transferring model is improved by training with generated cases (the S2S case-base). It validates that we can enhance the performance of DL models by generating high-quality case-bases.

The CycleGAN model significantly performs poorly even lower than the unprocessed samples. It may be because the data set is not large enough, which make CycleGAN overfitting \cite{peng2022contour}. Besides, AE performs badly because it directly smooths the \textit{internal noise} when transfers the position related \textit{environmental noise}. It will also lead to overfitting problems because the \textit{internal noise} can show low-frequency noise patterns that are superimposed on \textit{environmental noise} and signal. The spurious peaks will mislead AE to transfer it to a wrong noise or signal pattern. This phenomenon does not occur in two other baselines, i.e., wavelet and DnCNN, demonstrating the satisfactory results of traditional signal noise processing methods and the DNN-based models' generalisation capability.

\section{Conclusions}\label{sec6}

In this paper, we propose a noise pattern transferring approach for learning noise pattern differences from S2S cases, which is the first work to enhance the learning effect of noise pattern transferring models by generating cases. The most novelty idea is that we generate an S2S case-base using the `feature extraction-matching' method to enhance the learning ability of the noise pattern transferring model. Experimental results on a COD concentration measurement task show that our GC-based noise pattern transferring model outperforms important baselines ranging from wavelet denoising, DNN, and generative models. 

The excellent performance of the generated cases on different CNPT models also validates that our model can be used as a plug-in to the noise processing systems. When existing cases cannot express enough features for noise processing systems, implementing reasonable case-generating methods will contribute to performing the relevant tasks. In future work, we will continue exploring principles and applications of generated cases in noise processing systems.

\bibliographystyle{splncs04}
\bibliography{sn-article}

\end{document}